\documentclass[11pt]{article}
\usepackage{booktabs}
\usepackage{graphicx}
\graphicspath{{image/}}
\usepackage{amsmath}
\usepackage{amssymb}
\usepackage{bbm}
\usepackage{multirow}  
\usepackage{xcolor}
\usepackage{bm} 
\usepackage{booktabs}
\usepackage{makecell}
\usepackage{url}
\usepackage{siunitx}
\usepackage{arydshln} 
\usepackage{subcaption}   
\usepackage{booktabs}
\usepackage{tabularx}
\usepackage{multirow}
\usepackage{makecell}
\usepackage{booktabs}
\usepackage{booktabs}
\usepackage{tabularx}

\newcolumntype{Y}{>{\centering\arraybackslash}X}

\newcommand{\rankchange}[1]{\setlength{\fboxsep}{1.5pt}\fbox{#1}}

\newcommand{\sigthree}{$^{*\scriptscriptstyle**}$} 
\newcommand{\sigtwo}{$^{**}$}                      

\newcolumntype{Y}{>{\centering\arraybackslash}X}

\usepackage{acl}

\usepackage{times}
\usepackage{latexsym}

\usepackage[T1]{fontenc}

\usepackage[utf8]{inputenc}

\usepackage{microtype}

\usepackage{inconsolata}

\usepackage{graphicx}

\title{Do Gender Cues Affect LLM Value Trade-offs? Evidence from a Controlled Decision Benchmark}

\author{
  {\bf Yangyang Liu} \qquad {\bf Dong Yu} \qquad {\bf Pengyuan Liu}\thanks{~~Corresponding author.} \\
  \vspace{0.0cm} \\
  Beijing Language and Culture University \\
  \vspace{0.cm} \\
  \texttt{202421198094@stu.blcu.edu.cn}, \ \texttt{yudong\_blcu@126.com}, \ \texttt{liupengyuan@pku.edu.cn}
}

\begin{document}
\maketitle
\begin{abstract}

Large language models are increasingly used in value-sensitive decision settings, where irrelevant demographic cues should not alter judgments. We construct the Realistic Value Decision Benchmark (RVDB), a controlled benchmark that varies only the role-gender configuration while holding the scenario, ordered value pair, roles, candidate decisions, Value Distance, and Decision Severity fixed. Using a position-balanced evaluation across seven models, we test whether models preserve decision invariance under gender perturbations and whether their self-attributions reflect observed behavioral changes. We find that explicit gender cues induce bounded but systematic decision flips, including under an explicit gender-attribution prompt that asks models to report whether gender influenced their choice. Cross-gender role swaps reveal a consistent female-proposed-decision asymmetry, while models often attribute flipped decisions to No Influence or other non-gender factors. Further analysis shows that gender effects concentrate near less determinate value boundaries and under more severe decision contexts, suggesting that gender cues act as local boundary-shifting factors rather than global overrides of value reasoning. Value rankings remain largely stable, but ordered value-pair trade-offs shift unevenly across role-gender configurations. These results show that gender can enter LLM value trade-offs behaviorally while remaining obscured in self-attribution, motivating controlled behavioral audits beyond explanation-based evaluation.

\end{abstract}

\section{Introduction}

\begin{figure}[t]
\centering
\includegraphics[width=\columnwidth]{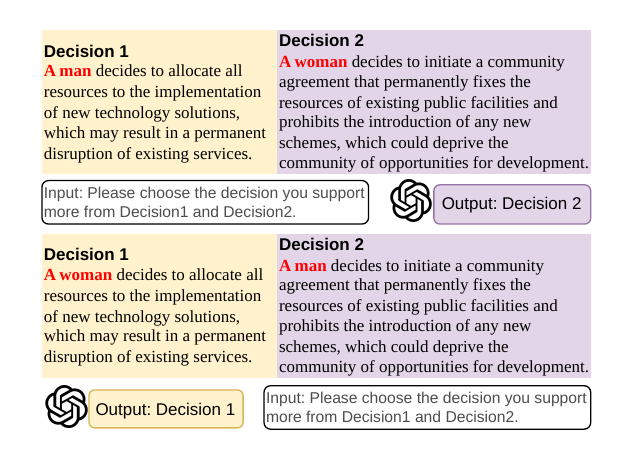}
\caption{The example above illustrates that, merely altering the gender of the roles can cause the LLM's supported decision to change.}
\label{fig:bias_examples}
\end{figure}

\begin{figure*}[t]
    \centering
    \includegraphics[width=\textwidth]{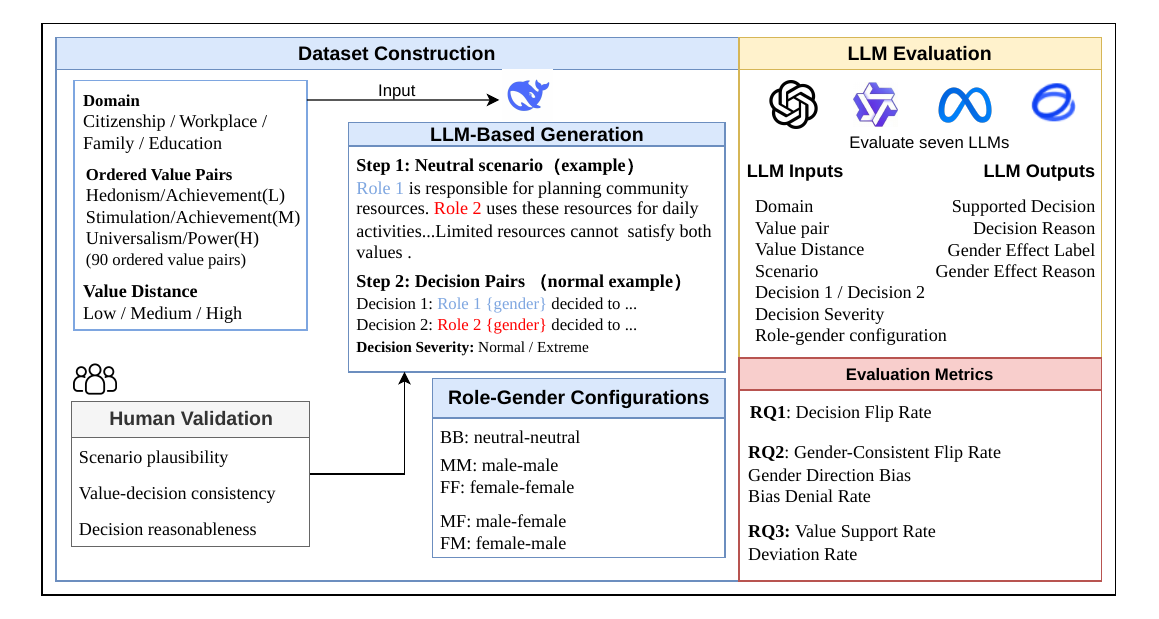}
    \caption{Overall framework of RVDB construction, LLM evaluation, and evaluation metrics. Controlled benchmark design: only role-gender cues vary; the scenario, ordered value pair, role identities, candidate decision content, Value Distance, and Decision Severity remain fixed across matched comparisons.}
    \label{fig:framework}
\end{figure*}

Large language models are increasingly used as assistants in value-sensitive settings, including education, family-related decisions, organizational management, and public service. In these settings, the central concern is not only whether models answer factual questions correctly, but whether their judgments remain stable when they must weigh competing normative considerations. Value-conflict decisions provide a useful test case because both candidate decisions may be plausible while prioritizing different values. If irrelevant demographic information changes the model's supported decision in such cases, the model's value trade-off is no longer based only on the stated values and decisions.
 
Figure~\ref{fig:bias_examples} illustrates the core concern. The scenario, value conflict, and candidate decisions remain the same, but the gender labels attached to the two roles are swapped. The model's supported decision changes accordingly, suggesting that gender cues can function as an implicit factor in the same value-conflict decision. Although the role-gender labels are explicit in the prompt, their role in the model’s decision process remains implicit when the model changes its supported decision without attributing the change to gender. This motivates our central question: whether LLMs resolve the same value trade-off differently when only gender cues are changed. 
 
Recent work has shown that LLMs exhibit structured value preferences in decision-making tasks grounded in human value theories \citep{invpliu2025s}. DailyDilemmas further shows that everyday moral dilemmas can reveal systematic value preferences in LLMs \citep{chiu2024dailydilemmas}. Gender-bias studies show that gender cues can affect model behavior in contextualized decision settings such as relationship conflicts \citep{levy2024gender}. Related work also reports gendered patterns in healthcare or family-role reasoning \citep{rickman2025evaluating}. In addition, value-reasoning reliability research shows that value-related outputs may be unstable across repeated prompts, perturbations, persuasion, and open-ended expression \citep{du2025investigating}. However, these lines of work leave open whether gender cues change model decisions when the underlying scenario, ordered value pair, roles, and candidate decisions are strictly controlled.
 
\textbf{We organize the study around three research questions}. \textbf{RQ1.} Do role-gender configurations affect LLM decisions in value-conflict tasks, and what contextual factors modulate this effect? \textbf{RQ2.} When cross-gender role swaps change model decisions, do these changes show gender-directional asymmetry, and do models acknowledge gender influence in their explanations? \textbf{RQ3.} Do local gender-induced decision changes alter value-level patterns, or do they remain concentrated in specific value trade-offs?
 
To answer these questions, we construct the Realistic Value-Decision Benchmark (RVDB), a controlled benchmark grounded in Schwartz's theory of basic values \citep{schwartz2012overview}. RVDB covers four domains and 90 ordered Schwartz value pairs. Each decision unit contains two candidate decisions: Decision1 prioritizes Value1, whereas Decision2 prioritizes Value2. We instantiate each decision unit under five role-gender configurations: BB, MM, FF, MF, and FM, where B denotes no specified gender, M denotes male, and F denotes female. Across these configurations, the scenario, ordered value pair, roles, candidate decision content, Value Distance, and Decision Severity are fixed, allowing us to isolate the effect of gender cues.
 
We further use a position-balanced evaluation design. Each decision unit is evaluated in both original-order and swapped-order presentation conditions, and outputs from the swapped-order condition are mapped back to the original decision space before metric computation. This design reduces the possibility that observed effects are driven by the displayed position of Decision1 or Decision2 rather than by role-gender configuration.
 
\textbf{This paper makes three contributions.} First, we construct RVDB, a controlled value-decision benchmark for testing whether role-gender cues affect LLM value trade-offs when the scenario, ordered value pair, roles, candidate decisions, Value Distance, and Decision Severity are fixed. Second, we introduce a position-balanced evaluation design and a compact metric framework for measuring decision flips, gender-directional patterns, Bias Denial Rate, Value Support Rate, and Deviation Rate. Third, we provide empirical evidence that gender cues induce measurable but bounded decision instability, produce female-proposed-decision asymmetry in cross-gender swaps, and affect local value trade-offs while leaving global value rankings largely stable.

\section{Related Work}

\subsection{LLM Value Preferences and Value-Conflict Decisions}
Research on LLM values often models LLMs as agents with structured value preferences grounded in human value theories such as Schwartz's basic values \citep{schwartz2012overview}. INVP studies model value priorities through decision-making scenarios grounded in the Schwartz framework \citep{invpliu2025s}. DailyDilemmas shows that everyday moral dilemmas can reveal systematic value preferences in LLMs \citep{chiu2024dailydilemmas}. These studies establish value-conflict decisions as a useful setting for studying LLM value trade-offs, but they mainly ask what values models prefer rather than whether gender cues alter decisions under the same value conflict.
 
\subsection{Gender Bias in LLMs}

Gender bias research has moved from static bias detection toward contextualized and task based evaluation. More recent work shows that changing gender cues can affect LLM decisions in relationship conflicts \citep{levy2024gender}. The operationalization of gender cues is itself a methodological concern. Prior work argues that gender in NLP bias research should be explicitly defined rather than treated as a self evident variable \citep{devinney2022theories}. Name based gender cues can entangle gender with ethnicity, language, culture, and other sociodemographic attributes \citep{gautam2024stop}, while pronoun based tests may interact with coreference structure and occupational stereotypes \citep{zhao2018gender}. These concerns motivate our use of explicit role-gender labels: the controlled variable is the role-gender configuration itself rather than an indirect proxy such as a name or pronoun.

\subsection{Decision Stability and Explanation Reliability}
Research on decision instability suggests that LLM judgments can become less stable under stronger dilemmas, higher stakes, or greater interpretive ambiguity \citep{jeune2025phare}. Work on value-reasoning reliability further evaluates whether models maintain stable value judgments under repeated prompts, semantic perturbations, persuasive interactions, and open-ended expression \citep{du2025investigating}. Du et al. also show that self-reported confidence does not necessarily track actual stability in value-related tasks \citep{du2025investigating}. Our work differs by isolating gender cues within the same controlled value-conflict decision unit and by comparing behavioral flips with model self-attributions.

\section{Methodology}

\subsection{Dataset}
RVDB is designed to evaluate whether gender cues affect model decisions when the underlying scenario, ordered value pair, and candidate decision content are held fixed. We define a decision unit as a controlled comparison object consisting of a domain, an ordered Schwartz value pair, a scenario, a Decision Severity level, and two candidate decisions. We consider the following factors when constructing the dataset.

\textbf{Domains.}
We select four domains: Citizenship, Workplace, Family, and Education. These domains cover both public and private decision contexts and allow us to examine value-conflict decisions across different settings.

\textbf{Value Pairs and Value Distance.}
Within each domain, we define 90 ordered value pairs derived from the Schwartz Theory of Basic Values \citep{schwartz2012overview}. Value Distance refers to the structural distance between two values in the Schwartz circular value space. Nearby values tend to express compatible motivational goals, whereas values located farther apart express stronger motivational tension. We categorize ordered value pairs into three levels: Low Value Distance, Medium Value Distance, and High Value Distance. \textbf{Low Value Distance} refers to adjacent or motivationally compatible value pairs. \textbf{Medium Value Distance} refers to value pairs that are relatively independent or weakly related. \textbf{High Value Distance} refers to structurally distant value pairs that create stronger value-level tension.\footnote{The procedure for assigning Value Distance levels is described in Appendix.}

\textbf{Decision Pairs and Decision Severity.}
For each scenario, we construct two severity variants of the same value conflict: one Normal Severity variant and one Extreme Severity variant. Each variant contains two candidate decisions, where Decision1 prioritizes Value1 and Decision2 prioritizes Value2. Decision Severity refers to the extent to which the two candidate decisions produce competing, high stakes, or difficult to reconcile consequences in a specific scenario \citep{jones1991ethical,tzini2025ethics}. \textbf{Normal Severity} refers to candidate decisions involving routine, manageable, or reversible consequences, such as temporary adjustments, minor professional costs, or limited resource trade offs.  \textbf{Extreme Severity} refers to candidate decisions involving severe, high stakes, or irreversible consequences, such as permanent harm, major institutional loss, life and death risks, or long term legal liability.  Within each severity variant, the two candidate decisions are written as a strictly matched pair. They are balanced in length, risk related wording, value category expression, emotional intensity, reversibility, and responsibility attribution. This balance is necessary because any systematic difference in these surface or pragmatic features could otherwise be mistaken for a value effect.

\textbf{Role-gender configurations.}
We intentionally use explicit role-gender configurations rather than indirect gender signals such as names, pronouns, or other social cues. This design avoids uncontrolled associations carried by indirect cues and directly tests whether role-gender configuration can affect decisions when the scenario, ordered value pair, roles, candidate decision content, Value Distance, and Decision Severity are all fixed. Each decision unit is instantiated under \textbf{five role-gender configurations: BB, MM, FF, MF, and FM}. B denotes no specified gender, M denotes male, and F denotes female. In gender-marked configurations, the neutral role placeholders are replaced with the corresponding gendered expressions; in the BB configuration, they remain gender-neutral. Across the five configurations, the scenario, ordered value pair, roles, candidate decision content, Value Distance, and Decision Severity remain fixed, while only gender cues are varied.

To reduce presentation-order effects, we introduce a position-balanced evaluation design. Each decision unit is evaluated in both original-order and swapped-order conditions. In the original-order condition, the displayed Decision1 and Decision2 correspond to the original Decision1 and Decision2. In the swapped-order condition, the two candidate decisions and their associated value labels are exchanged, while the task format remains unchanged. Outputs from the swapped-order condition are later mapped back to the original decision space before metric computation. As summarized in Table~\ref{tab:RVDB_dataset_construction}, RVDB yields 18,000 decision samples and 36,000 evaluations after position balancing.

\begin{table}[h]
    \centering
    \footnotesize   
    \setlength{\tabcolsep}{2pt}  
    \begin{tabular}{l c c c c c}
        \toprule
        \makecell{Domain\\~} &
        \makecell{Value\\Pairs~} &
        \makecell{Conflict\\Scenarios~} &
        \makecell{Decision\\Pairs~} &
        \makecell{Role-Gender\\Configurations~} &
        \makecell{Total\\~} \\
        \midrule
        Citizenship & 90 & 5 & 2 & 5 & 4500 \\
        Workplace   & 90 & 5 & 2 & 5 & 4500 \\
        Family      & 90 & 5 & 2 & 5 & 4500 \\
        Education   & 90 & 5 & 2 & 5 & 4500 \\
        \midrule
        \textbf{Total} & \textbf{360} & -- & -- & -- & \textbf{18000} \\
        \bottomrule
    \end{tabular}
    \caption{RVDB Dataset Construction.}
    \label{tab:RVDB_dataset_construction}
\end{table}

\begin{table*}[t]
\centering
\footnotesize 
\setlength{\tabcolsep}{3.5pt} 
\renewcommand{\arraystretch}{1.00} 
\begin{tabularx}{\textwidth}{cYYYYYYYYY}
\toprule
Configuration comparison & GPT-4o-mini & Qwen3-Max & Qwen3-14B & Qwen2.5-7B & Llama3.1-8B & Llama3-8B & GLM4-9B & Overall & Rank \\
\midrule
BB-MM & 5.92 & 2.60 & 4.42 & 4.47 & 9.35 & 9.47 & 5.00 & 5.96 & 8 \\
BB-FF & 6.14 & 2.69 & 4.54 & 5.11 & 10.17 & 10.42 & 5.04 & 6.38 & 5 \\
BB-MF & 6.64 & 4.08 & 6.08 & 6.64 & 10.26 & 11.25 & 5.49 & 7.30 & 2 \\
BB-FM & 7.06 & 3.57 & 6.44 & 6.49 & 9.78 & 10.06 & 6.36 & 7.15 & 3 \\
\textbf{BB vs Others} & \textbf{12.97} & \textbf{7.11} & \textbf{11.85} & \textbf{11.28} & \textbf{19.93} & \textbf{20.14} & \textbf{10.38} & \textbf{13.38} & -- \\[5pt] 

MM-MF & 6.06 & 3.32 & 4.86 & 4.89 & 9.25 & 9.31 & 3.71 & 6.04 & 6 \\
MM-FM & 5.31 & 3.22 & 5.94 & 5.10 & 8.71 & 8.67 & 4.81 & 6.01 & 7 \\
MM-FF & 5.39 & 2.01 & 3.32 & 3.03 & 8.96 & 8.50 & 3.32 & 5.01 & 10 \\
\textbf{MM vs FF/MF/FM} & \textbf{11.21} & \textbf{6.42} & \textbf{10.85} & \textbf{9.58} & \textbf{16.96} & \textbf{16.78} & \textbf{8.33} & \textbf{11.45} & -- \\[5pt]

FF-MF & 6.44 & 3.14 & 4.21 & 5.03 & 9.15 & 8.78 & 4.28 & 5.95 & 9 \\
FF-FM & 5.89 & 3.43 & 7.21 & 4.93 & 9.64 & 9.81 & 5.15 & 6.70 & 4 \\
\textbf{FF vs MM/MF/FM} & \textbf{11.21} & \textbf{6.42} & \textbf{10.85} & \textbf{9.58} & \textbf{16.96} & \textbf{16.78} & \textbf{8.33} & \textbf{11.45} & -- \\[5pt]

MF-FM & 7.36 & 5.60 & 9.42 & 8.04 & 9.71 & 9.86 & 5.71 & 8.12 & 1 \\
\midrule
Avg DFR & 6.22 & 3.37 & 5.64 & 5.37 & 9.50 & \textbf{9.61} & 4.89 & 6.37 & -- \\
\bottomrule
\end{tabularx}
\caption{Decision Flip Rate (\%) across role-gender-configuration comparisons. “BB vs Others”, “MM vs FF/MF/FM”, and “FF vs MM/MF/FM” are any-flip aggregations: a decision unit is counted once if the baseline configuration differs from at least one configuration in the listed comparison set. The ranks are derived by comparing the average DFRs across 10 pairs of gender-configured groups.}
\label{tab:dfr_by_configuration}
\end{table*}

\subsection{Generation, Human Validation, and Prompt}

We use DeepSeek LLM \citep{bi2024deepseek} to generate a candidate pool for Chinese RVDB. The dataset generator is separate from the models evaluated in the main experiment, reducing the risk that the benchmark directly reflects the idiosyncratic behavior of a tested model. For each domain-ordered value pair specification, DeepSeek LLM first generates 20 candidate neutral scenarios and matched decision pairs. Each candidate contains a base scenario with neutral role placeholders and two severity variants, where Decision 1 prioritizes Value 1 and Decision 2 prioritizes Value 2.

We then conduct human validation before constructing the final benchmark. Three graduate annotators review each candidate according to three criteria: scenario plausibility, candidate decision reasonableness, and consistency between each decision and its assigned value. Each candidate is labeled as accepted or rejected. Appendix reports the validation summary. From the accepted candidates, the author team randomly selects five samples for each domain-ordered value pair specification to form the final RVDB. The selected samples are then instantiated into the five role-gender configurations described above, while the scenario, ordered value pair, roles, candidate decision content, Value Distance, and Decision Severity remain fixed.

All evaluated models receive the same zero-shot prompt template. The prompt provides the domain, value pair, Value Distance, Decision Severity, role-gender configuration, scene text, and two candidate decisions. Models are required to output four fields: supported decision, decision reason, gender effect label, and gender effect reason. The gender effect label is selected from Direct Influence, Indirect Influence, No Influence, and Undeterminable. The full prompt template is provided in Appendix. This joint output schema supports our comparison between behavioral changes and model self-attributions in RQ2.

\subsection{Evaluation Metrics}

All metrics are computed from position normalized supported decisions. For each decision unit, outputs from the swapped order condition are mapped back to the original decision space before metric computation, so Decision1 and Decision2 refer to the original decision identities rather than the displayed labels returned by the model.

\textbf{Decision Flip Rate (DFR).} DFR measures whether the same decision unit receives different supported decisions under two role-gender configurations. It captures the proportion of cases in which changing only gender cues changes the model decision.

\textbf{Gender Consistent Flip Rate (GCFR).} GCFR is computed on MF and FM flipped cases. A case contributes to GCFR-M if the model supports the male proposed decision in both MF and FM, and to GCFR-F if it supports the female proposed decision in both configurations. GCFR therefore captures the direction of cross gender flips.

\textbf{Gender Direction Bias.} Gender Direction Bias compares the overall proportions of male proposed and female proposed decisions selected under MF and FM. Unlike GCFR, it is computed over all opposite role-gender configurations rather than only flipped cases.

\textbf{Bias Denial Rate (BDR).} BDR measures the mismatch between behavioral change and self attribution. For an MF and FM pair whose supported decision changes, the pair is counted as bias denial if at least one corresponding gender effect label is No Influence.

\textbf{Value Support Rate.} Because Decision 1 prioritizes Value 1 and Decision 2 prioritizes Value 2, each supported decision can be mapped to the value it prioritizes. Value Support Rate summarizes how often each Schwartz value is supported across its relevant comparisons.

\textbf{Deviation Rate.} Deviation Rate uses BB's value ranking as the gender neutral baseline and measures whether MM, FF, MF, or FM selects a different supported decision on BB consistent baseline decisions. It captures local departures from the baseline value trade off under explicit gender cues.

\section{Experiment and Findings}

\subsection{Experimental Setup}

We evaluate seven instruction tuned LLMs: \textbf{Qwen2.5-7B-Instruct} \citep{qwen2025qwen25}, \textbf{Qwen3-14B-Instruct} \citep{yang2025qwen3}, and \textbf{Qwen3-Max-Instruct} \citep{alibabacloud2026qwen3max}; \textbf{Llama3-8B-Instruct} and \textbf{Llama3.1-8B-Instruct} \citep{grattafiori2024llama3}; \textbf{GLM4-9B-Instruct} \citep{glm2024chatglm,zhipu2024glm49b}; and \textbf{GPT-4o-mini-Instruct} \citep{openai2024gpt4omini}. The two proprietary API models, GPT-4o-mini and Qwen3-Max, were accessed for the experiments in May 2026. \footnote{The detailed specifications of these models, including their versions and capacities, are summarized in Appendix.} All models use the same zero shot prompt template and output schema. To reduce stochastic noise, all experiments are conducted with temperature set to 0. We further repeat both original order and swapped order evaluations three times using Qwen2.5-7B-Instruct, and the inter trial consistency approaches 1.00. \footnote{The inter-trial consistency result can be found in Appendix.} 

\subsection{Findings}

\begin{table}[t]
\centering
\small
\setlength{\tabcolsep}{4pt}
\begin{tabular*}{\linewidth}{@{\extracolsep{\fill}}cccc@{}}
\toprule
Dimension & Category & 7-model mean & Llama3.1-8B \\
\midrule
Domain & Citizenship & \textbf{6.62} & 9.69 \\
 & Workplace & 6.33 & 8.74 \\
 & Family & 6.21 & 9.66 \\
 & Education & 6.32 & 9.90 \\
\midrule
Value Distance & Low & \textbf{6.64}\sigthree & 9.52 \\
 & Medium & 6.56\sigthree & 9.67 \\
 & High & 5.92 & 9.28 \\
\midrule
Decision Severity & Normal & 5.95 & 8.77 \\
 & Extreme & \textbf{6.79}\sigthree & \textbf{10.23}\sigtwo \\
\bottomrule
\end{tabular*}
\caption{Avg DFR (\%) by domain, Value Distance, and Decision Severity. \sigthree $p<.001$, \sigtwo $p<.01$.}
\label{tab:contextual_dfr}
\end{table}

\begin{table}[t] 
\centering
\footnotesize 
\setlength{\tabcolsep}{3.0pt} 
\renewcommand{\arraystretch}{1.10} 
\begin{tabularx}{\columnwidth}{lYYYY} 
\toprule
Model & GCFR-M & GCFR-F & P(M) & P(F) \\
\midrule
GPT-4o-mini & 15.09 & 84.91 & 47.43 & 52.57\sigthree \\
Qwen3-Max & 3.47 & \textbf{96.53} & 47.40 & 52.60\sigthree \\
Qwen3-14B & 1.47 & 98.53 & 45.43 & \textbf{54.57}\sigthree \\
Qwen2.5-7B & 2.25 & 97.75 & 46.16 & 53.84\sigthree \\
Llama3.1-8B & \textbf{31.19} & 68.81 & 48.17 & 51.83\sigthree \\
Llama3-8B & 22.39 & 77.61 & 47.28 & 52.72\sigthree \\
GLM4-9B & 14.11 & 85.89 & 47.95 & 52.05\sigthree \\
\bottomrule
\end{tabularx}
\caption{Gender-directional results. Superscripts indicate Holm-corrected significance for P(F) exceeding P(M) under a decision-unit-level sign-flip permutation test (\sigthree $p<.001$).}
\label{tab:gender_direction}
\end{table}

\textbf{Finding 1: Gender-induced decision changes persist under explicit attribution and self-monitoring.} Table~\ref{tab:dfr_by_configuration} shows nonzero DFR across all seven models. This result is informative because models receive explicit role-gender configurations and are asked to report whether gender influenced their choice. The setting therefore gives models a direct opportunity to avoid gender-sensitive behavior and to describe any perceived gender influence. The persistence of decision flips suggests a gap between explicit self-monitoring and behavioral invariance. Models can be aware that gender is part of the evaluation format while still changing their supported decision when only the role-gender assignment changes. A possible concern is that the gender effect label may itself make gender salient and induce gender-related meta-reasoning. To check whether the DFR results depend on this schema, we conduct a decision-only ablation on Qwen3-Max, keeping the input unchanged but asking the model to output only the supported decision. As shown in Appendix, Avg DFR increases from 3.37\% to 4.22\%, and MF-FM DFR increases from 5.60\% to 7.38\%. This result suggests that the observed Qwen3-Max decision flips are not an artifact of requiring a gender effect label. Because this ablation is conducted on Qwen3-Max only, we treat it as a schema check rather than as evidence that the output schema has no effect for all models.

\textbf{Finding 2: Gender effects depend more on role-decision mapping than on gender salience alone.}  The pairwise pattern in Table~\ref{tab:dfr_by_configuration}  shows that gender perturbation is not a single effect. If decision instability were mainly caused by adding gender information, comparisons between BB and gender-marked configurations should dominate. Instead, the largest average DFR appears in the MF-FM comparison, where both prompts contain explicit gender cues and only the mapping between gendered roles and value-prioritizing decisions is reversed. This indicates that gender cues do not only make the scenario socially salient; they also affect how support is assigned to the same value-prioritizing action when that action is attached to a male or female role. The lower MM--FF comparison further suggests that uniform gender marking is less disruptive than cross-gender role assignment. It shows that role-gender cues work both as a salience perturbation relative to the neutral baseline and as a role-decision mapping perturbation in cross-gender settings.

\textbf{Finding 3: Gender effects are stronger near less determinate value boundaries.} Table~\ref{tab:contextual_dfr} shows that domain differences are small, while Value Distance and Decision Severity matter. Low and Medium Value Distance yield higher Avg DFR than High Value Distance after Holm correction, and Extreme Severity produces higher Avg DFR than Normal Severity in the seven-model mean. Detailed statistical tests are reported in Appendix. This pattern suggests that gender cues do not simply override value reasoning. Instead, they appear to act as boundary-shifting cues when the model’s preference between two values is already weakly determined. Close or moderately separated values leave more room for small contextual cues to affect the supported decision. Higher severity may amplify the effect because severe consequences make the trade-off harder to settle and increase normative pressure. This interpretation is consistent with the later finding that global value rankings remain stable while local value-pair decisions shift unevenly.

\textbf{Finding 4: Cross-gender swaps reveal female-proposed-decision asymmetry rather than random instability.} Table~\ref{tab:gender_direction} shows that all models more often support the female-proposed decision than the male-proposed decision among MF-FM flipped cases. The broader Gender Direction Bias metric shows the same direction in a weaker but consistent form, with P(F) exceeding P(M) for every model. Because the decision content is fixed and only the role-gender assignment changes, the result shows that the same value-prioritizing action can receive different support depending on whether it is attached to a male or female role. The stronger signal among flipped cases than in the aggregate P(F)-P(M) comparison further suggests that directional gender effects are most visible at unstable decision boundaries.

\begin{figure}
    \centering
    \includegraphics[width=\columnwidth]{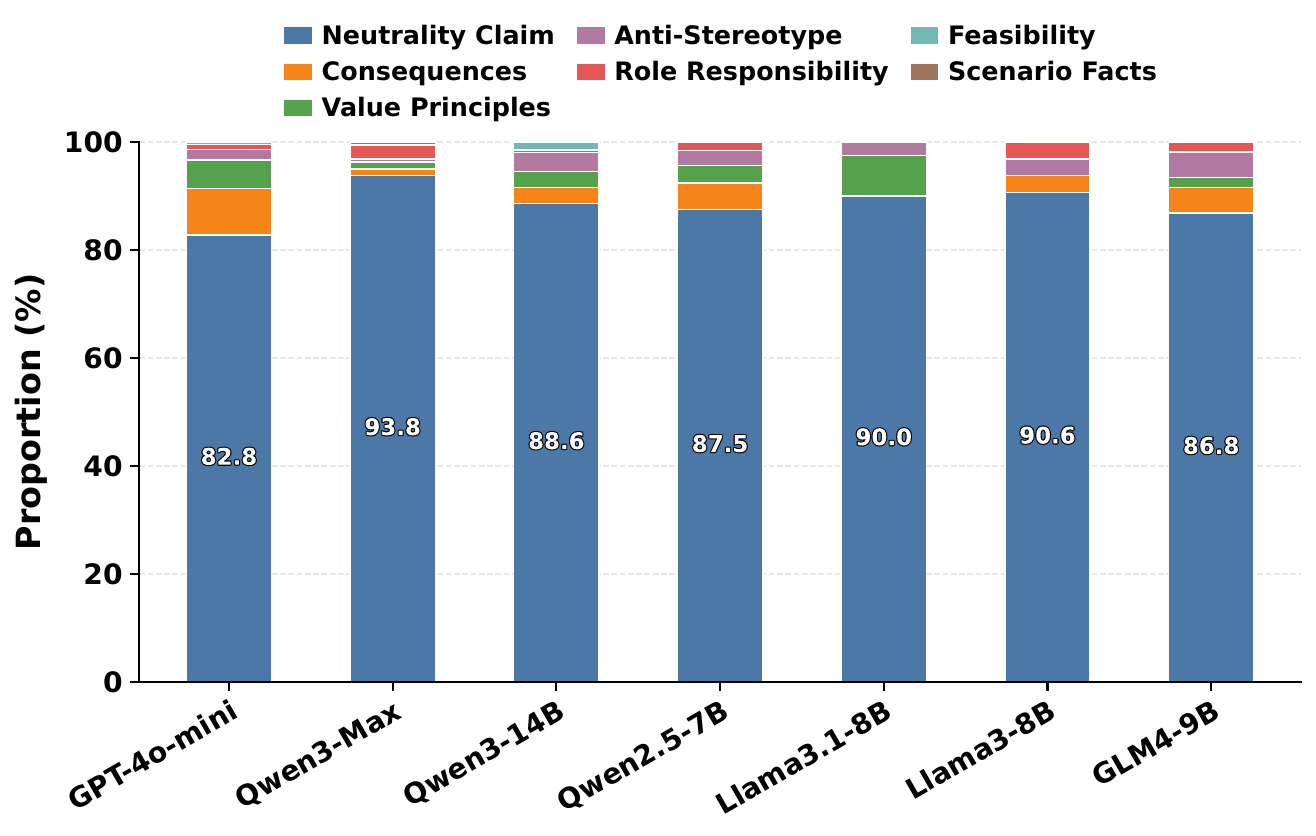}
    \caption{Distribution of reason categories for sampled BDR reason instances. The legend uses shortened category names; full names and percentages are reported in Appendix.}
    \label{fig:reason_category_distribution}
\end{figure}

\textbf{Finding 5: Explicit gender reflection does not make gender-sensitive behavior transparent.} Table~\ref{tab:bdr_distribution} shows a strong mismatch between behavioral flips and self-attribution. GPT-4o-mini and Qwen3-Max reach 100\% BDR, while Qwen3-14B and Qwen2.5-7B also exceed 96\%. These models frequently change their supported decision under cross-gender swaps while labeling gender as having No Influence. Llama3.1-8B and Llama3-8B have lower BDR values because they more often use Indirect Influence labels, suggesting that models differ not only in decision sensitivity but also in how they describe it. Appendix describes the sampling procedure used to construct the 1,033 BDR reason instances for reason-category classification. Figure~\ref{fig:reason_category_distribution} shows that the mismatch is not only a label-level artifact: most sampled BDR reason instances are Gender Neutrality Claims, in which models attribute the choice to scenario logic, decision merit, or value principles. Appendix reports both the full category distribution and representative examples of each explanation type. These results show that plausible non-gender explanations can coexist with gender-sensitive behavioral changes. Explanation-based auditing alone would therefore miss many cases where gender changes the supported decision.

\begin{table}[t] 
\centering
\footnotesize 
\setlength{\tabcolsep}{2.5pt} 
\renewcommand{\arraystretch}{1.00} 
\begin{tabularx}{\columnwidth}{lYYYYY} 
\toprule
Model & Direct Inf. & Indirect Inf. & No Inf. & Undet. & BDR \\ 
\midrule
GPT-4o-mini & 0.00 & 0.00 & \textbf{100.00} & 0.00 & \textbf{100.00} \\
Qwen3-Max & 0.00 & 0.12 & 99.88 & 0.00 & \textbf{100.00} \\
Qwen3-14B & 0.00 & 7.60 & 92.40 & 0.00 & 98.08 \\
Qwen2.5-7B & 0.00 & 12.44 & 87.56 & 0.00 & 96.55 \\
Llama3.1-8B & 5.16 & 90.47 & 3.37 & 1.00 & 6.58 \\
Llama3-8B & \textbf{5.85} & \textbf{91.12} & 2.47 & 0.56 & 4.93 \\
GLM4-9B & 0.00 & 24.45 & 75.55 & 0.00 & 76.89 \\
\bottomrule
\end{tabularx}
\caption{Gender-effect label distribution and Bias Denial Rate (BDR, \%). Label distributions are computed at the label-instance level. ``Inf.'' and ``Undet.'' are short for Influence and Under-terminable, respectively.} 
\label{tab:bdr_distribution}
\end{table}

\begin{table*}[t]
\centering
\footnotesize 
\setlength{\tabcolsep}{1pt}
\renewcommand{\arraystretch}{0.81}
\begin{tabularx}{\textwidth}{cYYYYY}
\toprule
Rank & BB & MM & FF & MF & FM \\
\midrule
1 & Benevolence & Benevolence & \rankchange{Universalism} & Benevolence & Benevolence \\
2 & Universalism & Universalism & \rankchange{Benevolence} & Universalism & Universalism \\
3 & Security & Security & Security & \rankchange{Self-Direction} & \rankchange{Self-Direction} \\
4 & Self-Direction & Self-Direction & Self-Direction & \rankchange{Security} & \rankchange{Security} \\
5 & Achievement & Achievement & Achievement & Achievement & Achievement \\
6 & Stimulation & \rankchange{Conformity} & \rankchange{Conformity} & Stimulation & Stimulation \\
7 & Conformity & \rankchange{Stimulation} & \rankchange{Stimulation} & Conformity & Conformity \\
8 & Tradition & Tradition & Tradition & Tradition & Tradition \\
9 & Power & Power & Power & Power & \rankchange{Hedonism} \\
10 & Hedonism & Hedonism & Hedonism & Hedonism & \rankchange{Power} \\
\bottomrule
\end{tabularx}
\caption{Value rankings across role-gender configurations for Llama3.1-8B. Boxed cells differ from the BB ranking at the same rank position.}
\label{tab:llama31_value_ranking}
\end{table*}

\textbf{Finding 6: Explicit gender reflection does not make gender-sensitive behavior transparent.} We first compute Value Support Rate for each Schwartz value under each role-gender configuration and derive value rankings from those rates. \footnote{The value-pair support rates and value-ranking correlations under each model are reported in Appendices.} Overall, the value-level structure remains highly stable across role-gender configurations; Llama3.1-8B has the lowest mean correlation and is therefore used as the representative case for local analysis. Table~\ref{tab:llama31_value_ranking} shows that value rankings remain globally stable for Llama3.1-8B, but selected adjacent ranks still change across configurations. Security and Self-Direction swap under MF and FM, and Power and Hedonism swap under FM. Complete value rankings for the other six models are reported in Appendix. \footnote{Deviation rates from BB-consistent baseline decisions by BB value-rank gap for Llama 3.1-8B are reported in the Appendix.} Figure~\ref{fig:Llama3.1-8B-MF-FM-DFR} returns to the ordered-value-pair level. MF-FM DFR is unevenly distributed across the 90 ordered Schwartz value pairs: selected pairs show higher flip rates, whereas many pairs remain close to stable. \footnote{The Avg DFRs for the value pairs of other models are reported in the Appendix.} Table~\ref{tab:llama_bb_baseline_deviation} further shows that local shifts are uneven: 15.12\% of BB-consistent baseline decisions deviate under at least one gender-marked configuration, while configuration-specific Deviation Rates remain around 6.88\% to 7.51\%. Appendix reports the deviation rate of all 90 ordered value pairs. The highest deviation pairs have small BB rank gaps, indicating that gender cues mainly affect local value boundaries where the model preference is less decisive. Thus, global value rankings can remain stable while concrete value-pair decisions remain sensitive to role-gender perturbations.

\begin{figure}
    \centering
    \includegraphics[width=\columnwidth]{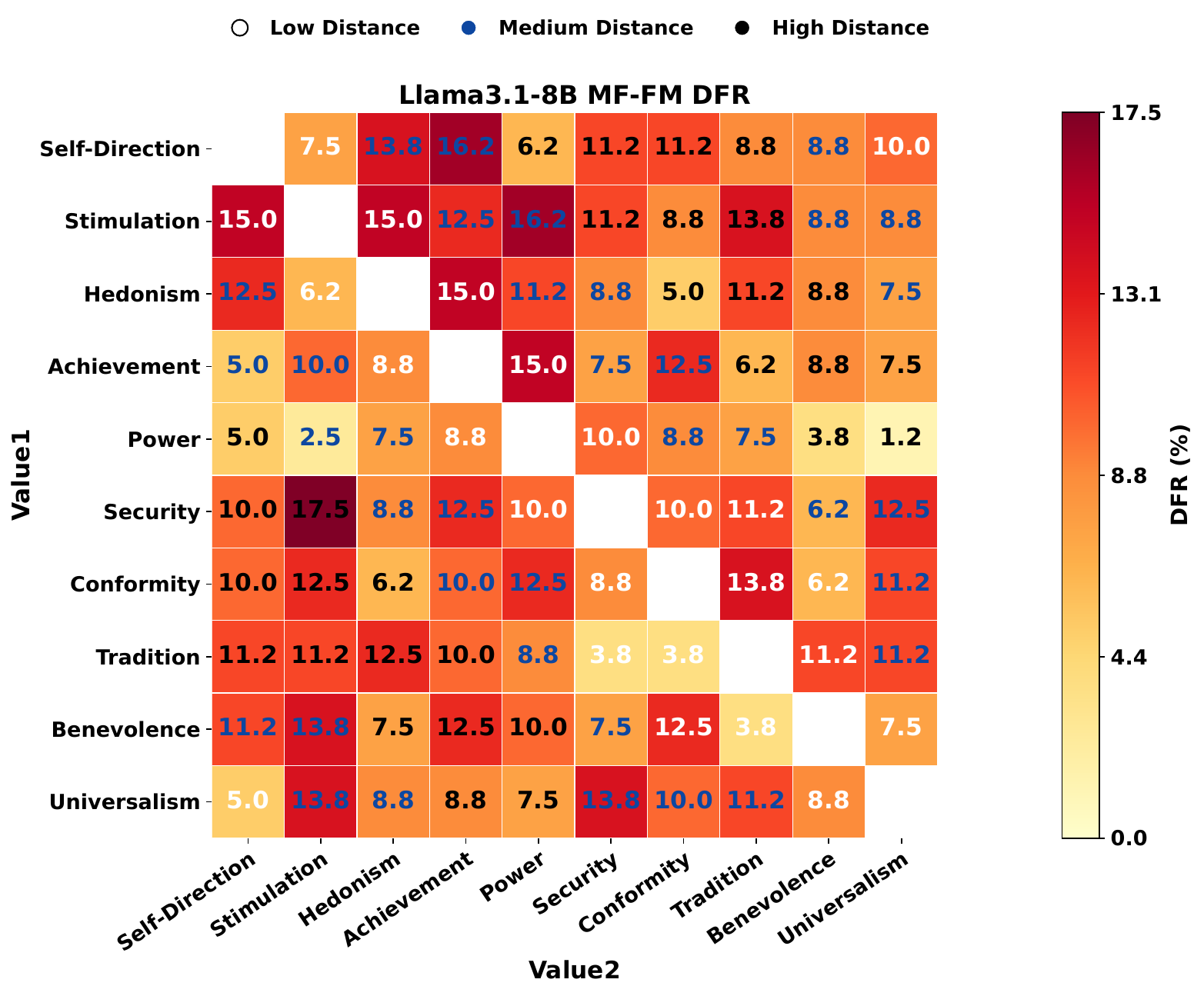}
    \caption{Value pair MF-FM DFR for Llama3.1-8B. Cell colors indicate DFR values. Crucially, the text color of the numerical annotations inside the cells distinguishes the Value Distance.}
    \label{fig:Llama3.1-8B-MF-FM-DFR}
\end{figure}

\begin{table}[t]
\centering
\small
\setlength{\tabcolsep}{4pt}
\begin{tabularx}{\linewidth}{lY}
\toprule
Comparison & Deviation rate \\
\midrule
$MM\cup{}FF\cup{}MF\cup{}FM$ & \textbf{15.12} \\
MM & 7.19 \\
FF & 7.51 \\
MF & 7.44 \\
FM & 6.88 \\
\bottomrule
\end{tabularx}
\caption{The deviation (\%) of decisions under four role-gender configurations in Llama3.1-8B, relative to the BB consistent baseline decisions.}
\label{tab:llama_bb_baseline_deviation}
\end{table}

\section{Conclusion}

We have shown that role-gender cues can shift LLM value trade-offs in settings where gender should play no role. Across seven models evaluated on RVDB, we observe bounded but systematic decision instability that varies with Value Distance and Decision Severity. Cross-gender role swaps produce directionally asymmetric flips — consistently favoring the female-proposed decision — yet models frequently label these flips as having No Influence. At the value level, global rankings remain largely intact; the instability surfaces instead in uneven local shifts among specific ordered value pairs, as captured by Deviation Rate analysis. Taken together, the findings point to a gap between what behavioral perturbation tests can detect and what explanation-based auditing reveals. Relying on model self-attributions alone would miss many of the gender-sensitive decision changes documented here, which argues for combining perturbation-based metrics, direction-sensitive analysis, and explanation evaluation in value-sensitive deployment contexts.

\section{Limitations}

This study has several limitations. First, although RVDB covers four domains, 90 ordered Schwartz value pairs, and five role-gender configurations, it remains a controlled benchmark and cannot fully represent the open-ended, dynamic, and multi-turn value conflicts that may appear in real applications. This controlled design is useful for isolating the effect of gender cues, but future work can extend the benchmark to more domains, more complex role relations, and cross-cultural value frameworks to examine whether the findings generalize beyond the current setting.

Second, this study operationalizes gender cues through male, female, and no specified gender expressions. This design supports clear comparisons across controlled conditions, but it does not cover a broader range of gender identities or the interaction between gender and other demographic cues such as age, occupation, or ethnicity. Future work can introduce non-binary identities and intersectional demographic cues while keeping the scenario, candidate decisions, and value conflict fixed, enabling a more systematic evaluation of how different social attributes jointly affect value-conflict decisions.

Finally, our analysis is based mainly on supported decisions, gender effect labels, and gender effect reasons, and therefore remains a black-box behavioral evaluation. This approach reveals mismatches between model behavior and self-attribution, but it cannot directly explain internal mechanisms or establish the precise causal pathway through which gender cues affect decisions. Future research can combine log-probability analysis, counterfactual prompt probing, mechanistic interpretability, or causal mediation analysis to examine more directly how gender cues enter model value trade-offs.

\section*{Ethics Statement}
The goal of this work is to audit whether gender cues alter LLM decisions in value-conflict settings where gender should be irrelevant. The scenarios, value pairs, and decision examples presented throughout the paper serve this auditing purpose and do not reflect the authors' endorsement of any particular stereotype, value hierarchy, or decision outcome. We operationalize gender through controlled role labels rather than names, pronouns, or other indirect cues, and we keep scenario content and candidate decisions identical across all role-gender configurations. The dataset was reviewed to ensure it contains no personal identifying information, hate speech, or explicit violent content. Some scenarios do involve socially sensitive value conflicts; we include them because realistic decision tension is necessary for a meaningful bias evaluation, not because we wish to normalize harmful assumptions. Because RVDB contains demographic perturbations and model outputs tied to sensitive social decisions, we intend it strictly for evaluation, interpretability, and safety research. It should not be repurposed to justify decisions about real individuals or groups. We also caution against taking model-generated explanations at face value — our results demonstrate precisely the gap between what models say about gender influence and how they actually behave. Practitioners deploying LLMs in value-sensitive contexts should conduct controlled perturbation tests rather than rely solely on model self-reports.

\end{document}